%% file: root.tex
\documentclass[letterpaper, 10 pt, conference]{ieeeconf}  

\IEEEoverridecommandlockouts                              

\usepackage[utf8]{inputenc}
\usepackage[T1]{fontenc}

\usepackage{graphicx}
\usepackage{xcolor}
\usepackage{amsmath}
\usepackage{amsfonts}
\usepackage{array}
\usepackage{booktabs}
\usepackage{caption}
\usepackage{subcaption}
\usepackage{longtable}
\usepackage{multirow}
\usepackage{wrapfig}
\usepackage{float}
\usepackage{xspace}
\usepackage{stfloats}

\usepackage{algorithm}
\usepackage{algpseudocode}

\usepackage{hyperref}
\usepackage{url}
\usepackage{microtype}
\usepackage{nicefrac}
\usepackage{lipsum}
\usepackage{soul}
\usepackage{textcomp}
\usepackage{capt-of}

\makeatletter
\def\fnum@figure{Figure~\thefigure}
\def\fnum@table{Table~\thetable}
\makeatother

\usepackage[textsize=tiny]{todonotes}
\usepackage{tcolorbox}
\tcbuselibrary{listings,breakable}

\definecolor{cbgreen}{HTML}{1F7A1F}
\definecolor{cbred}{HTML}{B22222}
\definecolor{darkgreen}{RGB}{0,128,0}
\definecolor{darkred}{RGB}{139,0,0}

\newcommand{\methodname}{RAGu\xspace}
\newcommand{\methodfullname}{\underline{R}etrieval-\underline{A}ugmented \underline{G}\underline{u}idance\xspace}
\newcommand{\baselinename}{Omniguide\xspace}

\newcommand{\srci}[2]{%
  \shortstack[c]{%
    #1\\[-2pt]%
    {\tiny\textcolor{gray}{[#2]}}%
  }%
}

\newcommand{\bsrci}[2]{%
  \shortstack[c]{%
    \textbf{#1}\\[-2pt]%
    {\tiny\textcolor{gray}{[#2]}}%
  }%
}

\title{\LARGE \bf
Zero-Shot VLA Adaptation with Retrieved Guidance
}

\author{
\authorblockN{\textbf{Subin Kim \quad Anh-Quan Pham \quad Dinesh Jayaraman}}
\authorblockA{University of Pennsylvania\\
{\tt\small \{kim34, quanpham, dineshj\}@engineering.upenn.edu}}
}

\begin{document}

\maketitle
\thispagestyle{empty}
\pagestyle{empty}
\input{tables/main_figure}

\begin{abstract}
Pre-training large parametric ``Vision-Language-Action'' (VLA) models on large demonstration datasets has recently yielded promising results for general-purpose manipulation. However, these models still struggle under even small domain shifts relative to the training data, such as changes in camera viewpoint, background, object identity, or robot morphology. 
Motivated by the success of Retrieval-Augmented Generation (RAG) in language models, we pursue retrieval to boost VLA performance. Specifically, we seek to leverage retrieval from domain-mismatched, "in-the-wild" demonstration datasets like DROID to guide a VLA's behavior in a new target domain. To achieve this, we introduce \methodname, which overcomes robotics-specific challenges for RAG: 
(1) It incorporates simple geometric calibration to transform source demonstrations into the target setup and reduce cross-domain differences.
(2) It uses a coarse-to-fine retrieval strategy that selects, from a large pool of pre-warped candidates, the trajectory most compatible with the base VLA.
(3) It guides the VLA's flow-matching inference procedure rather than seek to retrain the base model or expand its limited context window. We evaluate \methodname across real-world, simulation, sim-to-real transfer, and cross-embodiment humanoid manipulation. We find it improves real-world VLA success rates from 20\% to 80\% and outperforms the closest-related baselines in sim. Project website: \href{https://raguvla.github.io/}{\texttt{raguvla.github.io}}
\end{abstract}

\section{INTRODUCTION}
\label{sec:intro}
Vision-language-action (VLA) models have emerged in recent years as a key step toward generalist robots.
However, their ability to generalize still falls short of foundation models in language, vision, and coding.
One reason is that, in robotics, both pretraining datasets and the models trained on them remain relatively small, with the latter constrained by the need for near-real-time execution. As a result, VLAs still struggle to generalize across variations in camera viewpoint, scene background, object identity, and robot embodiment.
 
For language models, ``retrieval-augmented generation (RAG)"~\cite{lewis2020retrieval} is a well-established technique to boost performance, especially of \textit{small} models. Given a new query, RAG fetches relevant documents from a large database, typically through nearest neighbor search in an appropriate feature space, and places them into the model's context to guide its performance on the target task. This augments the limited ``parametric knowledge'' of the small language model with ``non-parametric knowledge'' in the form of the retrieved examples. 

In this paper, we seek to develop a RAG-style mechanism to improve generalization in vision-language-action (VLA) models. We target \textbf{three key problems in implementing RAG for robotics}: \textbf{(1) Cross-Domain Generalization.} 
Robotic data suffers from severe distribution shifts in camera viewpoints, layouts, and embodiments, unlike the universal token space of LLMs.
For a given target scenario, even the best-matched trajectories in large-scale robotics datasets are still likely to exhibit such \textit{domain-shifts}.
Prior robotic retrieval and RAG-style methods therefore restrict retrieval pools to narrow, identical-embodiment datasets or rely on task-specific fine-tuning of policy networks~\cite{memmel2025strap, regent2025, sridhar2025ricl}, fundamentally limiting their overall utility.
(2) \textbf{Policy-Aware Retrieval.} Good retrievals must be \textit{policy-aware}. Proximity based on simple measures like geometry do not suffice for a robot policy to follow a retrieved trajectory smoothly and stably. 
\textbf{(3) Context-Efficient Adaptation.} VLAs have limited context windows that do not permit directly concatenating retrieved demonstration trajectories to current observations. 

We propose \methodfullname (\textbf{\methodname}), a zero-shot, training-free VLA adaptation framework depicted in Figure~\ref{fig:metaworld_droid},
that handles all three challenges:
\begin{enumerate}
  \item \textbf{Cross-Domain Generalization.} It handles domain shifts through explicit injection of geometric and semantic priors based on Cartesian spatial traces in source data.
  \item \textbf{Policy-Aware Retrieval.} It selects source trajectories for guidance based on \textit{compatibility with the VLA} rather than only geometric proximity, and this selection is scaled efficiently through offline pre-warping and lightweight online scoring.
  \item \textbf{Context-Efficient
Adaptation.} Finally, rather than concatenating retrieved examples to the VLA's input, it constructs a test-time guidance field to steer the VLA's outputs.
\end{enumerate}

\section{Related Work}

\textbf{Retrieval and In-Context Learning for VLAs.}
Recent methods adapt VLA models using retrieval and in-context examples. REGENT~\cite{regent2025}, RICL~\cite{sridhar2025ricl} and RA-VLA~\cite{jang2026ra} retrieve demonstrations to improve adaptation to new tasks or environments, while STRAP~\cite{memmel2025strap} retrieves relevant sub-trajectories to support test-time policy learning. 
Unlike our training-free framework,
these methods all require in-domain trajectories as source demonstrations to support parametric adaptation.

\textbf{Inference-Time Steering of Generative Policies.}
Works such as VLS~\cite{liu2026vls}, Omniguide~\cite{omniguide2026}, and DemoDiffusion~\cite{du2023demodiffusion} frame adaptation as inference-time control by using external energy fields to steer generative policies, similar to our \methodname. However, these methods typically assume human demonstrations
or depend on sparse, VLM-induced waypoints,
and they lack a mechanism to handle large kinematic and spatial mismatches when the guidance comes from structurally different sources. \methodname specifically targets this domain mismatch.

\textbf{Cross-Domain Generalization with Trace Spaces.}
Prior works abstract manipulation into 3D representations~\cite{molmoact2025, liang2026tether, zhang2025peek} or train unified models for cross-domain transfer~\cite{shi2025zeromimic, li2025novaflow}, yet both rely on parametric training. In contrast, \methodname requires no retraining, enabling frozen VLAs to leverage cross-domain trajectories beyond their original distribution.

\section{Background}
\label{sec:background}
\textbf{Inference-Time Steering for VLA Policies.}
At time-step \(t\), we model the VLA as a flow-matching policy that maps a noisy action representation \( A_0 \)
to a clean action chunk \( A_1 = [a_t, a_{t+1}, \cdots, a_{t+H-1}] \) over denoising time \( \tau \)
~\cite{omniguide2026, intelligence2025pi05}.
During inference time, the model approximately integrates a conditional vector field \( v(A_{\tau}, o) \):
\[
\frac{dA_{\tau}}{d\tau} = v(A_{\tau}, o),
A_0 \sim \mathcal{N}(0,I), A_{\tau} \in \mathbb{R}^{H \times d}, \tau \in [0,1] (1)
\]
where \(o\) is the current observation, \(H\) is the horizon length of each action chunk, and \(d\) is robot joint space dimension. To adapt at inference time without fine-tuning, we impose a differentiable
constraint \( y \) through an energy \( \mathcal{L}_y(A_{\tau}) = -\log p(y \mid A_{\tau}) \)
~\cite{omniguide2026}, yielding the guided velocity
\[
v_{\text{guided}}(A_{\tau}, o)
= v(A_{\tau}, o) - \gamma \nabla_{A_{\tau}}\mathcal{L}_y(A_{\tau}), \qquad (2)
\]
where \( \gamma \) controls guidance strength. This steers action generation at
inference time while leaving the policy parameters unchanged. We let \( y = [y_t, y_{t+1}, \ldots, y_{t+H-1}] \in \mathbb{R}^{H \times d} \) denote the desired trajectory over the horizon, and \( \mathcal{L}_y(A_{\tau}) = \tfrac{1}{2}\|A_{\tau} - y\|_2^2 \). Prior works have assumed \( y \) to be provided from human operator~\cite{omniguide2026} or an external planner~\cite{liu2026vls}.

\textbf{In-Domain Trajectory Warping.} 
Recent work such as Tether~\cite{liang2026tether} extract critical 3D waypoints \( W = \{w_1, \dots, w_T\} \) from source demonstrations and map them to target waypoints \( W' = \{w'_1, \dots, w'_T\} \) via visual correspondences, to handle novel but within-domain scene configurations. Intermediate actions \( a_t \) between \( w_i \) and \( w_{i+1} \) are then warped to \( a'_t = a_t + \big( (1 - \alpha)\,\Delta d_i + \alpha\,\Delta d_{i+1} \big) \), based on displacements \( \Delta d_i = w'_i - w_i \) and interpolation \( \alpha \in [0,1] \) to execute inverse kinematics.
While direct inverse kinematics for cross-domain warped paths yields unreliable, suboptimal execution, we overcome this by treating the warped paths solely as spatial priors, allowing the VLA's innate generalization to manage the closed-loop manipulation.

\section{Cross-Domain \methodfullname for Zero-Shot VLA Adaptation}
\label{sec:method}

As motivated in Section~\ref{sec:intro}, using retrieved examples from a large out-of-domain source dataset to assist a pre-trained VLA faces three challenges: cross-domain shifts in viewpoint, geometry, or embodiment; the candidate closest in trajectory space is not necessarily one the policy can follow; and the lack of context window for demonstration tokens.
As shown in Figure~\ref{fig:metaworld_droid}, \methodname first \emph{filters} semantically relevant demonstrations and \emph{warps} them into the target scene via semantic correspondences, turning cross-domain demonstrations into usable spatial priors. It then \emph{retrieves} the guiding trajectory via coarse offline dynamic time-warping (DTW)-based pruning followed by policy-compatible selection at inference time, avoiding control-loop overhead. Finally, it \emph{steers} the VLA's flow-matching denoising toward the chosen trajectory through an energy-based guidance field, leaving the policy's weights and context untouched.

\textbf{Semantic Task-Category Filtering.} 
To ensure alignment with the target task, we first curate an initial pool of structurally compatible trajectories from large-scale datasets. We achieve this by filtering unstructured real-world datasets~\cite{khazatsky2024droid} using semantic keywords for broad task categories that share similar kinematic profiles. For example, we query demonstrations with keyword \textit{"pick"} and \textit{"place"} included for the task, \textit{"putting towel on the plate"}.
See Fig~\ref{fig:metaworld_droid} (a). 

\textbf{Trajectory Warping Across Different Domains.}
Unlike in-domain applications, cross-domain environments frequently exhibit significant semantic shifts, often lacking identical object instances or shared categories. To overcome this domain gap, we replace strict in-domain visual matching from prior work \cite{liang2026tether} with extracting semantic bounding boxes for task-relevant entities in the target scene via Grounding DINO~\cite{liu2024grounding}, effectively localizing the corresponding target waypoints \( W' \). See Fig~\ref{fig:metaworld_droid} (b). 
We define correspondence here not by object similarity, but at the level of functional roles it plays within a task.
Once these semantic correspondences are established, the remainder of the trajectory warping process proceeds identically to the formulation described in Section \ref{sec:background}. 

\textbf{Primary Retrieval: Coarse-grained static filtering via DTW.}
For each task category, we designate a single reference trajectory by taking an arbitrary valid demonstration--typically the first entry--from the semantically filtered pool. This one-time reference requires no per-episode tuning or meticulous curation.
Using this trajectory as a reference, we then apply Dynamic Time Warping (DTW)~\cite{senin2008dynamic} once for every episode to prune the remaining candidates down to the top-\(n\) trajectories, $N$.
Consequently,
\methodname maintains \(n\) warped candidate trajectories
\(\{W^{(i)}_t\}_{i \in [N]}\) in the target frame, where
\(W^{(i)}_t = \{w^{(i)}_{t},\dots,w^{(i)}_{t+H-1}\}\) denotes the waypoint
sequence of candidate \(i\) at time step \(t\). 
Note that \methodname does not assume matched domains in the source data: Across all experiments, target scenes differ from retrieved demonstrations in viewpoint, appearance, object identity, and goal location, requiring only shared task categories. 

\input{tables/experiments_combined_figure}

\textbf{Secondary Retrieval: Fine-grained inference-time policy alignment.}
At time-step \(t \), we estimate the VLA's output action chunk \(\hat{A}_\tau\) referred from Section~\ref{sec:background} with \(\tau=1\),  and map to a Cartesian 
end-effector
path
\(X_t = \{x_t,\dots,x_{t+H-1}\}\) via forward kinematics applied to \(\hat{A}_\tau\). For candidate trajectory \(W^{(i)}_t\), we define
\[
\mathcal{C}_i(X_t) =
\left\| X_t -
\mathrm{seqMatch}\big(X_t, W^{(i)}_t\big) \right\|_2^2 \qquad (3)
\]
where \(\mathrm{seqMatch}\) is a monotone nearest-neighbor projection approach (details on website).
We select the trajectory whose
cost gradient has the smallest norm:
\[
i^\star = \arg\min_{i \in N}
\left\| \nabla_{A_{0}} \mathcal{C}_i(X_t) \right\|_2 \qquad (4)
\]
To understand this choice, observe that this gradient norm is an estimate of how much the VLA must be steered away from its original intent to
reduce \(\mathcal{C}_i\).
By selecting trajectories with small gradient values, we obtain reference paths that stay close to the VLA's learned priors over optimal actions.
We perform this selection once at the first denoising
step of each inference and keep the selected trajectory \(W^{(i^\star)}_t\) fixed
for the remaining denoising steps. 
This may be thought of as the last and ``finest'' stage of our multi-stage coarse-to-fine retrieval process for implementing ``in-the-loop'' near-real-time retrieval augmentation for VLAs.

\textbf{Inference-Time Steering with Action Denoising.}
Once \(i^\star\) is selected, we use the corresponding warped trajectory to instantiate the inference-time guidance energy introduced in Section~\ref{sec:background}:
\[
\mathcal{L}_{\text{warp}}(A_{\tau}) = \mathcal{C}_{i^\star}(X_t). \qquad (5)
\]
Substituting \(\mathcal{L}_{\text{warp}}\) for \( \mathcal{L}_y(A_{\tau})\) into Equation~(2) steers the denoising process toward the retrieved trajectory while preserving the VLA prior. Candidate retrieval is performed once at the beginning of inference, and the selected trajectory remains fixed throughout the remaining denoising steps.

\section{Experiments}
\label{sec:experiments}

Based on the three challenges introduced in Section~\ref{sec:intro}, our experiments study the following: (1) \textit{Context-Efficient Adaptation.} How can retrieved demonstrations guide a VLA without consuming its limited context window?
(2) \textit{Cross-Domain Generalization.} Can cross-domain demonstrations guide a frozen VLA in new domains?
(3) \textit{Policy-Aware Retrieval.} Does retrieving dynamically compatible trajectories improve VLA performance? 

\begin{figure*}[t]
    \centering
    \noindent\centerline{
    \includegraphics[width=1.0\textwidth]{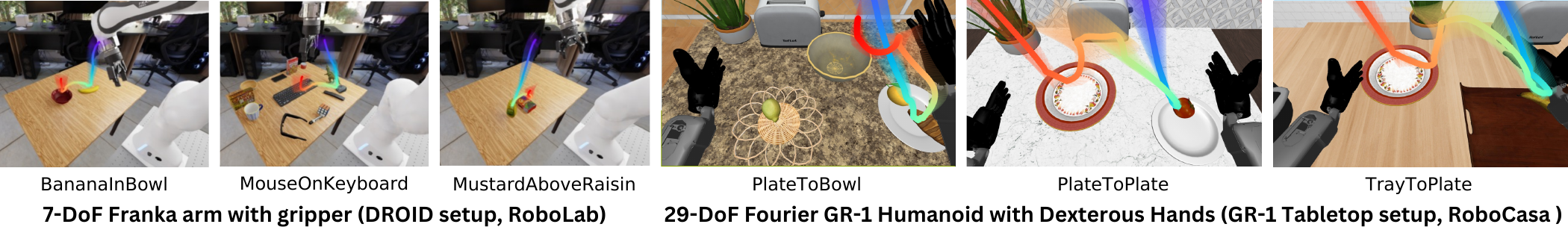}
    }
  \caption{\textbf{Real-to-Sim Cross-Embodiment Warping.} 
Real-world DROID demonstrations are warped to provide structural priors in simulated environments. The background streams visualize the distribution of candidate trajectories retrieved via coarse-grained DTW filtering.}
    \label{fig:edroid2sim_warping}
    \vspace{-0.8em}
\end{figure*}

\begin{figure*}[ht]
    \includegraphics[width=\textwidth]{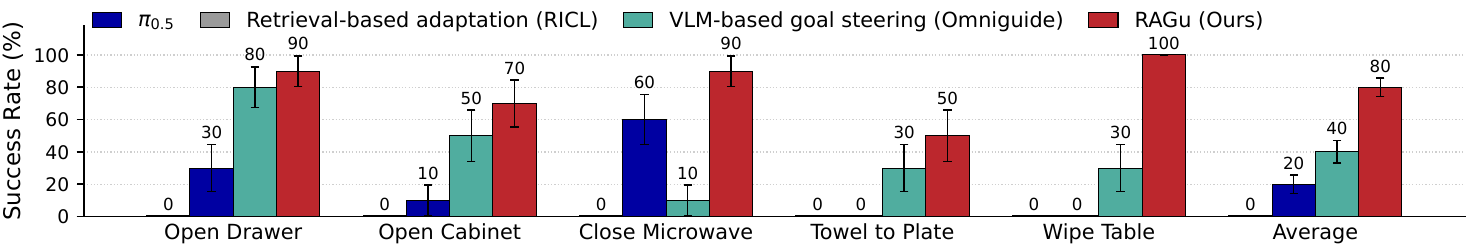}

    \centering
    \noindent\centerline{
    \includegraphics[width=1\textwidth]{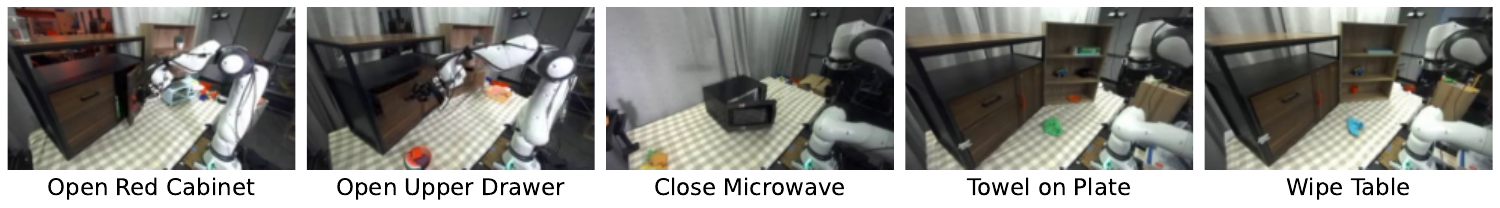}
    }
  \caption{\textbf{Real-World Success Rates.} Comparing \methodname against the base flow-matching policy (\(\pi_{0.5}\)), retrieval-based in-context learning baseline (RICL) and a VLM-based goal steering baseline (\baselinename). We evaluate the models across 5 diverse manipulation tasks: \texttt{Open Drawer}, \texttt{Open Cabinet}, \texttt{Close Microwave}, \texttt{Towel to Plate}, \& \texttt{Wipe Table}.}
    \label{fig:env_setup}
\vspace{-1.5em}
\end{figure*}

\subsection{Experimental Setup}
\textbf{Tasks \& Environments.} We evaluate four transfer settings using 10 rollouts for real-world experiments and 25 for simulation experiments. These include real-to-real (DROID~\cite{khazatsky2024droid} 7-DoF Franka, 5 tasks), real-to-sim (DROID to RoboLab~\cite{yang2026robolab}, 9 tasks), sim-to-real (MetaWorld~\cite{yu2019meta} Sawyer to real DROID, multi-stage \texttt{Banana-to-Plate} task), and cross-morphology (DROID to Fourier GR-1 humanoid in RoboCasa~\cite{robocasa2024}, 5 tabletop tasks~~\cite{nvidia2025gr00tn1openfoundation}), shown in Figure~\ref{fig:exp-combined}. DROID provides the source demonstrations for all except the sim-to-real setting, where it is publicly available and does not need additional data collection. Refer to Figure~\ref{fig:edroid2sim_warping} and the project website.

\textbf{Baselines.}
\textit{Retrieval-based in-context adaptation baseline:} 
We compare against
\textbf{RICL}~\cite{sridhar2025ricl}, which performs retrieval from exactly the same
pool of 100 warped DROID demonstrations as \methodname during test-time, and uses retrieved trajectories as input context for the VLA to infer actions. 
We use its released checkpoint, previously trained for in-context adaptation, with no test-time weight updates.
\textit{VLM-based goal steering baseline:} We also compare against \textbf{\baselinename}~\cite{omniguide2026} (real world) and \textbf{VLS}~\cite{liu2026vls} (simulation), a steering method that uses a Vision-Language Model to predict static goal object locations and defines an energy function around them. Compared to \methodname, these models replace retrieval and warping procedures with an external VLM, thereby representing approaches that condition on sparse waypoints or goal states rather than continuous trajectories.
\textit{Base VLA policy baseline:} \textbf{\(\pi_{0.5}\)-DROID}~\cite{intelligence2025pi05} (real world, Figure~\ref{fig:hardware_setup}) and \textbf{GR00T-N1.7} (simulation) were used, while both pre-trained and post-trained \textbf{GR00T-N1.5} open-sourced checkpoints~\cite{nvidia2025gr00tn1openfoundation} were used for GR-1 humanoid experiments as the frozen base policies. No weight updates occur during test time for all of the models in our experiments.

\textbf{Metrics, Protocol, and Measurements.}
Our primary metric is task success rate (Figures~\ref{fig:env_setup} and~\ref{fig:robolab_barplot};
Tables~\ref{tab:robolab_barplot}, \ref{tab:humanoid_results}, and~\ref{tab:robolab_trajectory_scaling_main}) over 10 rollouts per task in the
real world and 25 in simulation. To
characterize execution beyond binary success, we report three end-effector trajectory metrics (Table~\ref{tab:selection_ablation} and~\ref{tab:traj_metrics_avg}) including EE SPARC
(spectral arc length~\cite{balasubramanian2011robust}), a unitless smoothness measure in which higher value means smoother trajectory. We also report inference latency as mean wall-clock
time per episode (Table~\ref{tab:humanoid_wallclock_avg}). 
Figures~\ref{fig:wipe_traj}, \ref{fig:edroid2sim_warping}, 
\ref{fig:hardware_setup},
and \ref{fig:robolab_failure_mode}
provide qualitative visualizations. Detailed values and rollout videos are on the project website. In the rest of this section, we analyze these raw measurements and qualitative results, seeking answers to our key questions.
\subsection{Context-Efficient Adaptation}
\label{sec:rq1}
First, we investigate whether retrieved demonstrations can improve a frozen VLA without expanding its context window or updating its weights. We compare \methodname against two dominant test-time baselines: in-context retrieval (prompt concatenation) and VLM goal steering (sparse waypoint conditioning).


\begin{table}[t]
    \centering
    \footnotesize
    \setlength{\tabcolsep}{2.5pt}
    \renewcommand{\arraystretch}{0.95}
    \begin{tabular*}{\columnwidth}{@{\extracolsep{\fill}}lcccccc@{}}
        \toprule
        \textbf{Task}
        & \shortstack[c]{\textbf{GR00T}}
        & \(\boldsymbol{\pi_{0.5}}\)
        & \textbf{RICL}
        & \textbf{VLS}
        & \textbf{\methodname} \\
        \midrule
        \texttt{BananaInBowl}
        & 0 & 4 & 0 & 12
        & \textbf{36} \\
        \texttt{MouseOnKeyboard}
        & 0 & 12 & 0 & 16
        & \textbf{24} \\
        \texttt{MustardAboveRaisin}
        & 4 & 0 & 4 & 8
        & \textbf{16} \\
        \midrule
        \textbf{Average}
        & 1.33 & 5.33 & 1.33 & 12
        & \textbf{25.33} \\
        \bottomrule
    \end{tabular*}
    \caption{
        \textbf{RoboLab Real-to-Sim Success Rates.}
        Comparing \methodname vs.~base VLA policies (\(\pi_{0.5}\), GR00T-N1.7), RiCL, and VLS.
        For fair comparison, RiCL and \methodname use the same pool of 100 demonstrations from DROID. VLS uses \(\pi_{0.5}\) as the base model. Also see Tab~\ref{tab:robolab_trajectory_scaling_main}
    }
\label{tab:robolab_barplot}
\vspace{-1.5em}
\end{table}

\begin{figure}[t]
    \centering
    \includegraphics[width=\columnwidth]{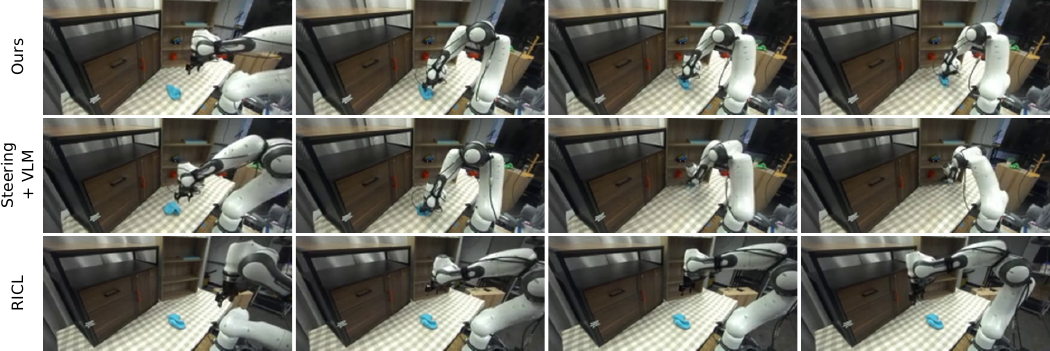}
    \caption{\textbf{Execution rollout comparison on the \texttt{Wipe Table} task.} \methodname continuously guides the end-effector along the table surface to execute a successful wipe. In contrast, \baselinename lifts the towel off the surface when transporting it toward the next discrete waypoint, failing to maintain contact with the table. RICL fails. (videos on website)}
    \label{fig:wipe_traj}
    \vspace{-1.5em}
\end{figure}

\textbf{\methodname vs.\ In-Context Retrieval.} See Figure~\ref{fig:env_setup} and Table~\ref{tab:robolab_barplot}.
Using the same pool of 100 warped DROID demonstrations, RICL~\cite{sridhar2025ricl} provides examples as in-context visual tokens. This yields zero-shot failure across all real-world tasks (0\%, Figure~\ref{fig:env_setup}) and 1\% in RoboLab (Table~\ref{tab:robolab_barplot}), falling below the unguided $\pi_{0.5}$ baseline (20\% and 5\%). In contrast, \methodname achieves 80\% and 25\%. In-context examples force the VLA to visually resolve domain gaps zero-shot, whereas trajectory guidance circumvents the context window to provide direct spatial priors the policy can immediately follow. Figure~\ref{fig:wipe_traj} shows an execution.

\textbf{\methodname vs.\ VLM Goal Steering.} Continuing with  Figure~\ref{fig:env_setup} and Table~\ref{tab:robolab_barplot}, Omniguide~\cite{omniguide2026} (real-world) and VLS~\cite{liu2026vls} (simulation) employ identical inference steering as \methodname, but steer toward sparse VLM goal points rather than dense trajectories. \methodname doubles their success in both settings (80\% vs.\ 40\% real-world; 25\% vs.\ 12\% RoboLab). This gap is pronounced in contact-rich tasks like \texttt{Wipe Table} (100\% vs.\ 30\%), where sparse goals cause Omniguide to break surface contact between waypoints (Figure~\ref{fig:wipe_traj}). Dense trajectories preserve approach and contact constraints that discrete goals omit.



\begin{table*}[t]
    \centering




    \footnotesize
    \setlength{\tabcolsep}{3pt}
    \resizebox{0.9\textwidth}{!}{%
    \begin{tabular}{l|ccccc|ccccc}
    \toprule
    \multirow{2}{*}{\textbf{Task}}
    & \multicolumn{5}{c|}{\textbf{GR00T-N1.5 (w/o FT)}}
    & \multicolumn{5}{c}{\textbf{GR00T-N1.5 (w/ FT)}} \\
    
    & \textbf{Base}
    & \textbf{Ours (1 traj)} & \textbf{$\Delta$}
    & \textbf{Ours (100 traj)} & \textbf{$\Delta$}
    & \textbf{Base}
    & \textbf{Ours (1 traj)} & \textbf{$\Delta$}
    & \textbf{Ours (100 traj)} & \textbf{$\Delta$} \\
    \midrule

        CuttingboardToPot
    & 4
    & 12 & \textcolor{darkgreen}{+8}
    & 12 & \textcolor{darkgreen}{+8}
    & 4
    & 8 & \textcolor{darkgreen}{+4}
    & 8 & \textcolor{darkgreen}{+4} \\
    
    PlacematToPlate
    & 48
    & 60 & \textcolor{darkgreen}{+12}
    & 72 & \textcolor{darkgreen}{+24}
    & 48
    & 72 & \textcolor{darkgreen}{+24}
    & 72 & \textcolor{darkgreen}{+24} \\
    
    PlateToBowl
    & 44
    & 44 & \textcolor{gray}{+0}
    & 56 & \textcolor{darkgreen}{+12}
    & 68
    & 72 & \textcolor{darkgreen}{+4}
    & 80 & \textcolor{darkgreen}{+12} \\
    
    PlateToPlate
    & 56
    & 44 & \textcolor{darkred}{-12}
    & 60 & \textcolor{darkgreen}{+4}
    & 20
    & 40 & \textcolor{darkgreen}{+20}
    & 36 & \textcolor{darkgreen}{+16} \\
    
    TrayToPlate
    & 16
    & 36 & \textcolor{darkgreen}{+20}
    & 36 & \textcolor{darkgreen}{+20}
    & 44
    & 40 & \textcolor{darkred}{-4}
    & 48 & \textcolor{darkgreen}{+4} \\
    \midrule
    
    \textbf{Average}
    & \textbf{33.6}
    & \textbf{39.2} & \textbf{\textcolor{darkgreen}{+5.6}}
    & \textbf{47.2} & \textbf{\textcolor{darkgreen}{+13.6}}
    & \textbf{36.8}
    & \textbf{46.4} & \textbf{\textcolor{darkgreen}{+9.6}}
    & \textbf{48.8} & \textbf{\textcolor{darkgreen}{+12}} \\

    \bottomrule
    \end{tabular}%
    }

    \caption{
    \textbf{Cross-embodiment humanoid manipulation with in-context trajectory guidance.}
    We report success rates on RoboCasa tasks using the base GR00T-N1.5 model and its fine-tuned variant. The $\Delta$ columns indicate absolute improvement over the corresponding unguided baseline. Refer to the project website for confidence intervals.}
    \label{tab:humanoid_results}
\vspace{-1.5em}

\end{table*}

\textbf{Varying Base Model Competence.}
As shown in Table~\ref{tab:humanoid_results}, \methodname improves both base and fine-tuned GR00T-N1.5 models (+13.6\% and +12.0\% with 100 trajectories). Rather than acting as a redundant replacement for parametric training, test-time trajectory guidance provides complementary gains across varying baseline policy capacities.

\subsection{Cross-Domain Generalization}
\label{sec:rq2}
We evaluate whether demonstrations from mismatched domains can guide a frozen VLA across shifts in viewpoint, scene, object identity, and morphology.

\begin{figure}[t]
    \centering
    \includegraphics[width=\columnwidth]{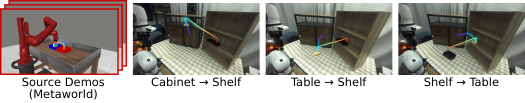}
    \includegraphics[width=\columnwidth]{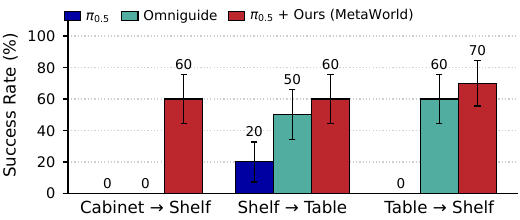}

    \caption{
       \textbf{Sim-to-real cross-embodiment transfer.}
        Success rates (\%) for each object-transfer
        setting. We warp 100 expert trajectories from the MetaWorld Sawyer robot (\texttt{bin-picking-v3}) to the real-world DROID setup.
    }
    \label{fig:robolab_barplot}
    \vspace{-1.5em}
\end{figure}

\textbf{Crossing Sim$\rightarrow$Real \& Real$\rightarrow$Sim Gaps.}
For real-to-sim, warping real-world DROID demonstrations into RoboLab raises $\pi_{0.5}$ success from 5.33\% to \methodname's 25.33\% (Table~\ref{tab:robolab_barplot}). For sim-to-real, warping 100 MetaWorld Sawyer trajectories to a real Franka setup enables multi-stage \texttt{Banana-to-Plate} execution (Figure~\ref{fig:robolab_barplot}). This gain peaks in Cabinet$\rightarrow$Shelf, where both $\pi_{0.5}$ and Omniguide fail (0\%) while \methodname achieves 60\%. In this out-of-distribution layout, operating on Cartesian traces makes guidance agnostic to whether sources are rendered or recorded.

\textbf{Embodiment Shifts.}
We test for morphological shift by using 7-DoF Franka DROID demonstrations to guide GR00T-N1.5 on a 29-DoF Fourier GR-1 humanoid (Table~\ref{tab:humanoid_results}), along with the results shown previously in Figure~\ref{fig:robolab_barplot} from Sawyer to Franka. With 100 candidates, average success rises by $+13.6\%$. This cross-embodiment transfer succeeds because guidance targets Cartesian wrist trajectories via differentiable forward kinematics, avoiding any explicit joint mapping, inverse kinematics, or trajectory replay. Refer to App.~\ref{sec:gr1} for implementation details.

\textbf{Visual and Environmental Diversity.}
We also highlight that all warped demonstrations are sourced from DROID~\cite{khazatsky2024droid}, which natively exhibits diversity in camera viewpoints, backgrounds, and object instances from our target setting.


\subsection{Policy-Aware
Retrieval}
\label{sec:rq3}
We now evaluate design choices within the trajectory selection rule of the secondary retrieval phase.

\begin{table}[t]
\centering
\scriptsize
\setlength{\tabcolsep}{2.5pt}
\renewcommand{\arraystretch}{1.05}
\resizebox{\columnwidth}{!}{
\begin{tabular}{@{}lccc@{}}
\toprule
& \textbf{EE SPARC} $\uparrow$
& \textbf{Path (m)} $\downarrow$
& \textbf{Speed (cm/s)} \\
\midrule
RAGu (1 traj)
& -9.93 & 4.61 & 8.20 \\

RAGu (min dist)
& -11.74 & 5.79 & 7.09 \\

\textbf{RAGu (min grad)}
& \textbf{-10.30} & \textbf{5.67} & \textbf{8.37} \\
\bottomrule
\end{tabular}
}
\caption{\textbf{Utility of policy-aware secondary retrieval.}
Trajectory metrics on RoboLab for three tasks in Table~\ref{tab:robolab_barplot}, 25 rollouts per task. We use the same pool of 100 warped DROID demonstrations, while only the rule that picks the guiding trajectory changes. \textit{1 traj} uses only a single heuristically selected trajectory. \textit{min dist} selects $\arg\min_i \mathcal{C}_i(X_t)$ (Eq.~3), and \textit{min grad} selects $\arg\min_i \|\nabla_{A_0}\mathcal{C}_i(X_t)\|_2$ (Eq.~4) from a 100 trajectory pool. Refer to \textit{Secondary Retrieval ablation} in the project website.}
\label{tab:selection_ablation}
\vspace{-1em}
\end{table}

\input{tables/secondary_retrieval_ablation}
\textbf{Secondary Retrieval Candidate Selection Rule.}
Table~\ref{tab:selection_ablation} evaluates Secondary Retrieval from a fixed pool of demonstration candidates. 
Replacing \methodname's gradient-norm selection (\textit{min grad}) with nearest-neighbor selection (\textit{min dist}) causes slower ($7.09$ vs.\ $8.37$\,cm/s) and less smooth (EE SPARC $-11.74$ vs.\ $-10.30$) motion at similar path lengths ($5.79$ vs.\ $5.67$\,m). 
A plausible explanation is that geometric proximity ignores the policy's learned priors, so \textit{min dist} may repeatedly pull the policy away from its intended motion and vice versa, whereas \textit{min grad} favors references that are more compatible with the policy.
In Table~\ref{tab:robolab_trajectory_scaling_main}, while a single reference achieves smoother motion, it drops success to 17.3\% compared to using 100 trajectories (25.3\%) due to limited scene coverage. 
\textit{min grad} retains the coverage of candidate pooling while recovering execution smoothness.

\input{tables/robolab_speed_ablation}
\input{tables/traj_num_ablation_main}
\textbf{Benefits of Larger Retrieval Pools.}
Expanding the candidate pool consistently enhances performance. Across nine RoboLab tasks (Table~\ref{tab:robolab_trajectory_scaling_main}), for a base unguided $\pi_{0.5}$ model averaging $23.6\%$ success,
\methodname boosts performance to $28.4\%$
with a single demonstration retrieval pool, and further to $36.4\%$ with 100 demos. Trajectory quality mirrors these gains (Table~\ref{tab:traj_metrics_avg}): the 100-demo pool produces the smoothest ($-9.32$) and most direct ($5.25$\,m) paths at comparable speeds, as denser candidate pools provide references that simultaneously match the target scene and align with the policy prior.

\begin{table}[t]
\centering
\scriptsize
\setlength{\tabcolsep}{2.5pt}
\renewcommand{\arraystretch}{1.05}
\resizebox{\columnwidth}{!}{%
\begin{tabular}{@{}lccc@{}}
\toprule
& \textbf{EE SPARC} $\uparrow$
& \textbf{Path length (m)} $\downarrow$
& \textbf{Speed (cm/s)} \\
\midrule
$\pi_{0.5}$
& -9.37 & 5.93 & 9.37 \\

$\pi_{0.5}$ + \methodname\ (1 traj)
& -9.35 & 5.37 & 8.67 \\

\textbf{$\pi_{0.5}$ + \methodname\ (100 traj)}
& \textbf{-9.32} & \textbf{5.25} & \textbf{8.89} \\
\bottomrule
\end{tabular}%
}
\caption{\textbf{Trajectory quality metrics across selected tasks in RoboLab.}  Using 100 trajectories reduces path length and increases smoothness at similar speeds, improving execution efficiency. We use the same selected tasks as Tab.~\ref{tab:robolab_trajectory_scaling_main}. Refer to \textit{Trajectory quality} in the project website.}
\label{tab:traj_metrics_avg}
\vspace{-1em}
\end{table}

\textbf{Inference Efficiency.}
Because cross-domain warping is handled offline and the $n$ candidates are evaluated only once at the first denoising step, selection incurs minimal computational overhead. As shown in Table~\ref{tab:humanoid_wallclock_avg}, mean wall-clock time per episode across five RoboCasa tasks remains within $6$\,s of the unguided baseline over 720-step episodes at 10\,Hz. Consequently, the retrieval pool can scale by two orders of magnitude without degrading closed-loop control frequencies.

\begin{table}[t]
    \centering
    \footnotesize
    \setlength{\tabcolsep}{2.2pt}
    \renewcommand{\arraystretch}{1.0}

    \begin{tabular*}{\columnwidth}{@{\extracolsep{\fill}}lccccc@{}}
    \toprule
    \textbf{GR00T-N1.5}
    & \textbf{Base}
    & \textbf{Ours-1}
    & \textbf{$\Delta$}
    & \textbf{Ours-100}
    & \textbf{$\Delta$} \\
    \midrule

    w/o FT
    & 34.1
    & 31.0
    & \textcolor{darkgreen}{-3.1}
    & 33.6
    & \textcolor{darkgreen}{-0.5} \\

    w/ FT
    & 26.8
    & 30.3
    & \textcolor{darkred}{+3.5}
    & 32.7
    & \textcolor{darkred}{+5.9} \\
    \bottomrule
    \end{tabular*}
    \caption{
    \textbf{Average inference wall-clock cost (s/episode).}
    We measure inference wall-clock cost of 5 tasks in Table~\ref{tab:humanoid_results} (max time steps/episode~$720$, control rate $10$\,Hz, $25$ episodes/task, \textcolor{darkgreen}{green}: faster,
    \textcolor{darkred}{red}: slower). The horizon, episode count, task set, and simulator configuration are held constant across all four methods.
    }
    \label{tab:humanoid_wallclock_avg}
\vspace{-1em}
\end{table}

\section{Conclusion}
\label{sec:conclusion}
We presented \methodfullname (\methodname), a zero-shot, training-free framework for adapting pretrained VLAs to OOD environments by warping cross-domain trajectories into the target geometric space and using them as guidance during flow-matching denoising. 
Across real-to-real, real-to-sim, sim-to-real, and arm-to-humanoid settings, \methodname improves frozen VLA policies without any parameter update, including transfer from simulated Sawyer demonstrations to multi-stage real-world Franka manipulation, while larger retrieval pools further improve robustness.

\textbf{Limitations.} \methodname assumes that the source database contains demonstrations that approximately match the target task category. The method is still not invariant to camera placement and occlusion, since semantic correspondence failures degrade 3D warping and corrupt the guidance signal. Finally, guidance can only reshape motion within the VLA's prior.

\section*{Acknowledgements}
This project was supported in part by DARPA TIAMAT HR0011249042, NSF SLES 2331783, Penn Engineering Futures Fund, NSF CAREER 2239301, ONR N00014-22-1-2677.

\section*{APPENDIX}

\textbf{Clarification of Confidence Intervals (Figure~\ref{fig:env_setup}, Figure~\ref{fig:robolab_barplot})} Error bars represent $\pm 1$ standard error of the binomial proportion, computed as $\text{SE} = \sqrt{p(1 - p) / n}$, where $p$ is the empirical success rate over $n$ rollouts ($n = 10$ per task, and $n = 50$ for the pooled average) and whiskers are clipped to $[0\%, 100\%]$. Because the standard error evaluates to zero when $p \in \{0, 1\}$, bars at 0\% and 100\% success do not display whiskers.

\textbf{Hyperparameter Selection and Guidance Sensitivity}
Table~\ref{tab:ragu_hyperparameters} shows the selected hyperparameters for experiments. The guidance strength parameter $\gamma$ balances policy priors with external trajectory constraints. Refer to Omniguide~\cite{omniguide2026} for comprehensive parameter-sweep curves on gradient-based flow steering.

\input{tables/hardware_setup}
\input{tables/traj_num_ablation_figure}

\begin{table}[t]
    \centering
    \footnotesize
    \setlength{\tabcolsep}{4pt}
    \renewcommand{\arraystretch}{1.1}
    \resizebox{\columnwidth}{!}{%
    \begin{tabular}{@{}lccc@{}}
    \toprule
    \textbf{Source} & \textbf{Success (\%)}
    & \textbf{EE SPARC} & \textbf{Speed (cm/s)} \\
    \midrule
    \(\pi_{0.5}\) - 25 rollouts/task
      & 14.0 & \(-9.04 \pm 3.91\) & \(10.3 \pm 2.1\) \\
    \(\pi_{0.5}\) - 10 rollouts/task~\cite{yang2026robolab}
      & 28.0 & \(-8.34 \pm 6.7\) & \(5.4 \pm 1.6\) \\
    \bottomrule
    \end{tabular}%
    }
    \vspace{-0.3em}
    \caption{\textbf{\(\pi_{0.5}\)-DROID sanity check on all 45 RoboLab
    tasks.} When re-evaluating the unguided base policy against the statistics reported in RoboLab~\cite{yang2026robolab}, we obtain a 14.0\% average success rate compared to their reported 28.0\%. We believe that the differences in performance may be due to the fact that the original paper evaluated on only 10 rollouts per task, so their reported success rates may be noisy. For fairness, we evaluate all methods in our tables are evaluated under identical conditions, ensuring that relative comparisons remain valid.}
    \label{tab:robolab_sanity}
\vspace{-1.5em}
\end{table}

\begin{table}[t]
\centering
\footnotesize
\setlength{\tabcolsep}{4pt}
\renewcommand{\arraystretch}{1.1}
\resizebox{\columnwidth}{!}{%
\begin{tabular}{@{}lccc@{}}
\toprule
& \textbf{DROID (Real)} & \textbf{DROID (RoboLab)} & \textbf{GR-1 (RoboCasa)} \\
\midrule
Top-\(N\)                  & 8        & 16       & 3        \\
Loss weight \(\lambda\)    & 1.0      & 1.0      & 1.0      \\
Guidance scale \(\gamma\)  & 0.2--2.0 & 0.2--1.0 & 0.4, 0.5 \\
\bottomrule
\end{tabular}%
}
\vspace{-0.3em}
\caption{\textbf{\methodname hyperparameters.} \(N\) is the number of
DTW-pruned trajectories retained; \(\lambda\) weights the trajectory
guidance loss; and \(\gamma\) is the denoising guidance scale.}
\label{tab:ragu_hyperparameters}
\vspace{-1.5em}
\end{table}

\input{tables/robolab_failure_mode}

\textbf{Failure Mode Analysis.}
For the RoboLab tasks, we estimate a 65.1\% warping success rate over 900 total attempts, with the remaining 34.9\% attributed to failures in Figure~\ref{fig:robolab_failure_mode}. This suggests that demo-side trajectory parsing errors are the dominant source of error, exceeding scene correspondence failures by roughly a factor of two. This indicates that improving source-demo quality control is likely to provide a larger benefit than further tuning the correspondence stage.

\textbf{Humanoid implementation details}
\label{sec:gr1}
GR00T N1.5 outputs normalized GR-1 joint-space action chunks, which are
denormalized and mapped to a Cartesian right-palm trajectory by
differentiable forward kinematics. The tracked palm point is an
\(8\,\mathrm{cm}\) offset along the negative local \(z\)-axis of the
right-hand frame. The predicted trajectory and warped DROID reference are
both expressed in the \texttt{mobilebase0\_support} and matched sequentially for
guidance. We use no inverse kinematics or direct trajectory replay. We add
two base-fixed \(60^\circ\) warp cameras at \([0.15,\pm0.05,0.5]\) to obtain
stable calibrated target-scene views.


\bibliographystyle{IEEEtran}
\bibliography{main}

\addtolength{\textheight}{-12cm}

\end{document}

%% file: tables/main_figure.tex
\begin{figure*}[t!]
    \centering
    \noindent\centerline{
    \includegraphics[width=1\textwidth]{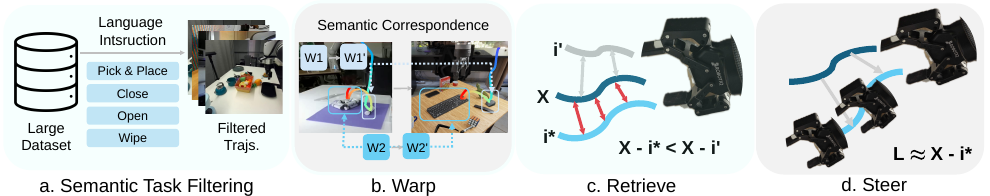}
    }
  \caption{\textbf{Overview of \methodname pipeline.}
    \textbf{(a)} We filter real-world datasets using semantic keywords for broad task categories with similar kinematic profiles, such as pick-and-place via language instruction labels in the datasets.
    \textbf{(b)}
    Task-relevant objects in the target scene are identified using semantic bounding box detection to extract target waypoints \( W' \), allowing the source trajectories to be smoothly adapted despite visual shifts.
    \textbf{(c)} Retrieval candidates $i$ are coarsely pruned to the top-$N$ using Dynamic Time Warping (DTW)-based matching. We then select the optimal trajectory $i^\star$ from pruned trajectories $i^{'} \in [N]$ that minimizes the guidance gradient norm ($||\nabla_{A_0} \mathcal{C}_{i^{*}}(X_t)||_2$), retrieving the demonstration best aligned with the VLA's expected Cartesian trajectory $X_t$.
    \textbf{(d)} At test time, the chosen reference $i^\star$ defines an energy field ($\mathcal{L}_{\text{warp}} \approx ||X_t - i^\star||_2$), steering diffusion denoising for zero-shot adaptation.
    }
    \label{fig:metaworld_droid}
    \vspace{-1.5em}
\end{figure*}

%% file: tables/experiments_combined_figure.tex
\begin{figure*}[t]
    \centering
    \noindent\centerline{
    \includegraphics[width=1.0\textwidth]{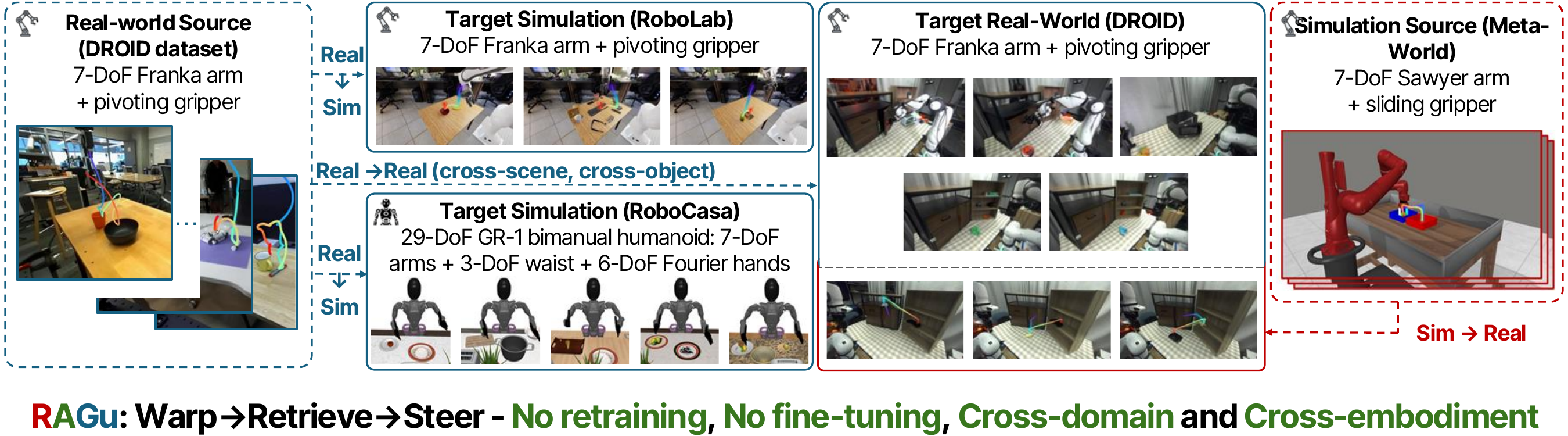}
    }
  \caption{\textbf{\methodname enables zero-shot adaptation across domains and embodiments.} \methodname transfers manipulation knowledge across four settings: \textbf{Real→Real, Real→Sim, Sim→Real} with different morphologies with the base VLA's weights completely frozen without any parameter update.
}
\vspace{-1.5em}
\label{fig:exp-combined}
\end{figure*}

%% file: tables/secondary_retrieval_ablation.tex



%% file: tables/traj_num_ablation_main.tex
\begin{table}[t]
\centering
\scriptsize
\setlength{\tabcolsep}{3pt}
\renewcommand{\arraystretch}{0.95}
\resizebox{\columnwidth}{!}{%
\begin{tabular}{@{}lcccccc@{}}
\toprule
\multirow{2}{*}{\textbf{Task}}
& \multirow{2}{*}{\textbf{GR00T}}
& \multirow{2}{*}{\(\boldsymbol{\pi_{0.5}}\)}
& \multicolumn{3}{c}{\(\boldsymbol{\pi_{0.5}}\) \textbf{+ \methodname} (\textbf{\# traj})}
& \multirow{2}{*}{\(\boldsymbol{\Delta}\)} \\
\cmidrule(lr){4-6}
& & & \textbf{1} & \textbf{10} & \textbf{100} & \\
\midrule

\texttt{BananaInBowl}
& 0 & 4 & 28 & 28 & \textbf{36}
& \textcolor{darkgreen}{+32} \\

\texttt{MouseOnKeyboard}
& 0 & 12 & 16 & 8 & \textbf{24}
& \textcolor{darkgreen}{+12} \\

\texttt{MustardAboveRaisin}
& 4 & 0 & 8 & 0 & \textbf{16}
& \textcolor{darkgreen}{+16} \\

\texttt{WoodSpatulaToBowl}
& 4 & 0 & 4 & 0 & \textbf{12}
& \textcolor{darkgreen}{+12} \\

\texttt{RubiksCubeOrBanana}
& 0 & 44 & 52 & 48 & \textbf{68}
& \textcolor{darkgreen}{+24} \\

\texttt{SauceBottlesCrate}
& 0 & 4 & 12 & \textbf{16} & 12
& \textcolor{darkgreen}{+8} \\

\texttt{TakeMeasuringSpoonOut}
& 0 & 36 & 32 & \textbf{44} & 36
& \textcolor{gray}{+0} \\

\texttt{RubiksCube}
& 0 & 52 & 20 & \textbf{68} & 64
& \textcolor{darkgreen}{+12} \\

\texttt{ToyInBin}
& 28 & 60 & \textbf{84} & 76 & 60
& \textcolor{gray}{+0} \\

\midrule

\textbf{Mean}
& 4.0 & 23.6 & 28.4 & 32.0 & \textbf{36.4}
& \textbf{\textcolor{darkgreen}{+12.9}} \\

\bottomrule
\end{tabular}%
}
\caption{\textbf{Scaling the retrieval pool on Real-to-Sim RoboLab.}
Task-wise success rates (\%, 25 rollouts per task) for GR00T-N1.7,
\(\pi_{0.5}\), and \(\pi_{0.5}\) + \methodname guided by 1, 10, or 100 warped
DROID demonstrations. \(\Delta\) is the absolute gain of the 100-trajectory
pool over \(\pi_{0.5}\); bold marks the best entry in each row. Expanding the
pool consistently helps: mean success rises from 23.6\% (unguided) to 28.4\%
with a single trajectory and 36.4\% with 100.
See Table~\ref{tab:robolab_trajectory_scaling} in the Appendix for the full
version.}
\label{tab:robolab_trajectory_scaling_main}
\vspace{-1em}
\end{table}

%% file: tables/hardware_setup.tex
\begin{figure}[H]
    \centering
    \begin{minipage}[c]{0.45\columnwidth}
        \centering
        \includegraphics[width=\linewidth]{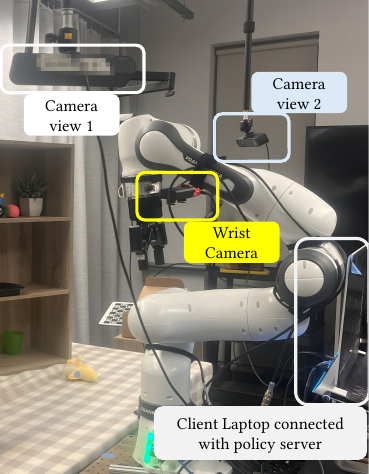}
    \end{minipage}%
    \hspace{0.03\columnwidth}%
    \begin{minipage}[c]{0.4\columnwidth}
        \captionsetup{justification=raggedright,singlelinecheck=false}
        \caption{\textbf{Hardware configuration for real-world evaluation.}
        We use the DROID~\cite{khazatsky2024droid} setup: a 7-DoF Franka
        Emika Panda arm, a Robotiq 2F-85 parallel-jaw gripper, a
        wrist-mounted ZED Mini RGB-D camera, and two side-mounted ZED 2
        stereo cameras.}
        \label{fig:hardware_setup}
    \end{minipage}
    \vspace{-0.8em}
\end{figure}

%% file: tables/traj_num_ablation_figure.tex
\providecommand{\srci}[2]{%
  \shortstack[c]{%
    #1\\[-2pt]%
    {\tiny\textcolor{gray}{[#2]}}%
  }%
}

\providecommand{\bsrci}[2]{%
  \shortstack[c]{%
    \textbf{#1}\\[-2pt]%
    {\tiny\textcolor{gray}{[#2]}}%
  }%
}

\begin{table*}[t]
\centering
\footnotesize
\setlength{\tabcolsep}{4pt}
\renewcommand{\arraystretch}{0.88}
\resizebox{\textwidth}{!}{%
\begin{tabular}{lcc*{5}{cc}}
\toprule
\multirow{2}{*}{\textbf{Task}}
& \multirow{2}{*}{\textbf{GROOT}}
& \multirow{2}{*}{\textbf{\(\pi_{0.5}\)}}
& \multicolumn{2}{c}{\textbf{\(\pi_{0.5}\) + Ours (1 traj)}}
& \multicolumn{2}{c}{\textbf{\(\pi_{0.5}\) + Ours (10 traj)}}
& \multicolumn{2}{c}{\textbf{\(\pi_{0.5}\) + Ours (30 traj)}}
& \multicolumn{2}{c}{\textbf{\(\pi_{0.5}\) + Ours (50 traj)}}
& \multicolumn{2}{c}{\textbf{\(\pi_{0.5}\) + Ours (100 traj)}} \\
\cmidrule(lr){4-5}
\cmidrule(lr){6-7}
\cmidrule(lr){8-9}
\cmidrule(lr){10-11}
\cmidrule(l){12-13}
& &
& \textbf{SR} & \textbf{\(\Delta\)}
& \textbf{SR} & \textbf{\(\Delta\)}
& \textbf{SR} & \textbf{\(\Delta\)}
& \textbf{SR} & \textbf{\(\Delta\)}
& \textbf{SR} & \textbf{\(\Delta\)} \\
\midrule

BananaInBowl
& \srci{0}{0.0, 13.3}
& \srci{4}{0.7, 19.5}
& \srci{28}{14.3, 47.6}
& \textcolor{darkgreen}{+24}
& \srci{28}{14.3, 47.6}
& \textcolor{darkgreen}{+24}
& \srci{28}{14.3, 47.6}
& \textcolor{darkgreen}{+24}
& \srci{24}{11.5, 43.4}
& \textcolor{darkgreen}{+20}
& \bsrci{36}{20.2, 55.5}
& \textbf{\textcolor{darkgreen}{+32}} \\

MouseOnKeyboard
& \srci{0}{0.0, 13.3}
& \srci{12}{4.2, 30.0}
& \srci{16}{6.4, 34.7}
& \textcolor{darkgreen}{+4}
& \srci{8}{2.2, 25.0}
& \textcolor{darkred}{-4}
& \srci{16}{6.4, 34.7}
& \textcolor{darkgreen}{+4}
& \srci{20}{8.9, 39.1}
& \textcolor{darkgreen}{+8}
& \bsrci{24}{11.5, 43.4}
& \textbf{\textcolor{darkgreen}{+12}} \\

MustardAboveRaisin
& \srci{4}{0.7, 19.5}
& \srci{0}{0.0, 13.3}
& \srci{8}{2.2, 25.0}
& \textcolor{darkgreen}{+8}
& \srci{0}{0.0, 13.3}
& \textcolor{gray}{+0}
& \srci{0}{0.0, 13.3}
& \textcolor{gray}{+0}
& \srci{0}{0.0, 13.3}
& \textcolor{gray}{+0}
& \bsrci{16}{6.4, 34.7}
& \textbf{\textcolor{darkgreen}{+16}} \\

WoodSpatulaToBowl
& \srci{4}{0.7, 19.5}
& \srci{0}{0.0, 13.3}
& \srci{4}{0.7, 19.5}
& \textcolor{darkgreen}{+4}
& \srci{0}{0.0, 13.3}
& \textcolor{gray}{+0}
& \srci{8}{2.2, 25.0}
& \textcolor{darkgreen}{+8}
& \srci{4}{0.7, 19.5}
& \textcolor{darkgreen}{+4}
& \bsrci{12}{4.2, 30.0}
& \textbf{\textcolor{darkgreen}{+12}} \\

RubiksCubeOrBanana
& \srci{0}{0.0, 13.3}
& \srci{44}{26.7, 62.9}
& \srci{52}{33.5, 70.0}
& \textcolor{darkgreen}{+8}
& \srci{48}{30.0, 66.5}
& \textcolor{darkgreen}{+4}
& \srci{52}{33.5, 70.0}
& \textcolor{darkgreen}{+8}
& \srci{56}{37.1, 73.3}
& \textcolor{darkgreen}{+12}
& \bsrci{68}{48.4, 82.8}
& \textbf{\textcolor{darkgreen}{+24}} \\

SauceBottlesCrate
& \srci{0}{0.0, 13.3}
& \srci{4}{0.7, 19.5}
& \srci{12}{4.2, 30.0}
& \textcolor{darkgreen}{+8}
& \srci{16}{6.4, 34.7}
& \textcolor{darkgreen}{+12}
& \srci{8}{2.2, 25.0}
& \textcolor{darkgreen}{+4}
& \bsrci{24}{11.5, 43.4}
& \textbf{\textcolor{darkgreen}{+20}}
& \srci{12}{4.2, 30.0}
& \textcolor{darkgreen}{+8} \\

TakeMeasuringSpoonOut
& \srci{0}{0.0, 13.3}
& \srci{36}{20.2, 55.5}
& \srci{32}{17.2, 51.6}
& \textcolor{darkred}{-4}
& \bsrci{44}{26.7, 62.9}
& \textbf{\textcolor{darkgreen}{+8}}
& \srci{32}{17.2, 51.6}
& \textcolor{darkred}{-4}
& \srci{36}{20.2, 55.5}
& \textcolor{gray}{+0}
& \srci{36}{20.2, 55.5}
& \textcolor{gray}{+0} \\

RubiksCube
& \srci{0}{0.0, 13.3}
& \srci{52}{33.5, 70.0}
& \srci{20}{8.9, 39.1}
& \textcolor{darkred}{-32}
& \bsrci{68}{48.4, 82.8}
& \textbf{\textcolor{darkgreen}{+16}}
& \srci{64}{44.5, 79.8}
& \textcolor{darkgreen}{+12}
& \srci{64}{44.5, 79.8}
& \textcolor{darkgreen}{+12}
& \srci{64}{44.5, 79.8}
& \textcolor{darkgreen}{+12} \\

ToyInBin
& \srci{28}{14.3, 47.6}
& \srci{60}{40.7, 76.6}
& \bsrci{84}{65.3, 93.6}
& \textbf{\textcolor{darkgreen}{+24}}
& \srci{76}{56.6, 88.5}
& \textcolor{darkgreen}{+16}
& \srci{64}{44.5, 79.8}
& \textcolor{darkgreen}{+4}
& \srci{64}{44.5, 79.8}
& \textcolor{darkgreen}{+4}
& \srci{60}{40.7, 76.6}
& \textcolor{gray}{+0} \\

\midrule

\textbf{Mean}
& \bsrci{4.0}{2.1, 7.4}
& \bsrci{23.6}{18.5, 29.5}
& \bsrci{28.4}{22.9, 34.7}
& \textbf{\textcolor{darkgreen}{+4.9}}
& \bsrci{32.0}{26.3, 38.4}
& \textbf{\textcolor{darkgreen}{+8.4}}
& \bsrci{30.2}{24.6, 36.5}
& \textbf{\textcolor{darkgreen}{+6.7}}
& \bsrci{32.4}{26.7, 38.8}
& \textbf{\textcolor{darkgreen}{+8.9}}
& \bsrci{36.4}{30.4, 42.9}
& \textbf{\textcolor{darkgreen}{+12.9}} \\

\bottomrule
\end{tabular}%
}
\caption{\textbf{Evaluation on Real-to-Sim RoboLab tasks with different numbers of guidance trajectories.}
Task-wise success rates (\%, 25 rollouts per task) across GROOT-N1.7,
\(\pi_{0.5}\), and \(\pi_{0.5}\) + \methodname guided by 1, 10, 30, 50,
or 100 warped DROID demonstrations. Values in brackets are 95\% Wilson
confidence intervals. Each \(\Delta\) column shows the absolute improvement
over \(\pi_{0.5}\). Multi-stage retrieval is required both to cover large
domain gaps and to meet real-time constraints. Shrinking the retrieval pool
to a single trajectory reduces mean RoboLab success from 36.4\% to 28.4\%.}
\label{tab:robolab_trajectory_scaling}
\vspace{-\baselineskip}
\end{table*}

%% file: tables/robolab_failure_mode.tex
\begin{figure}[ht]
    \centering
    \includegraphics[width=\columnwidth]{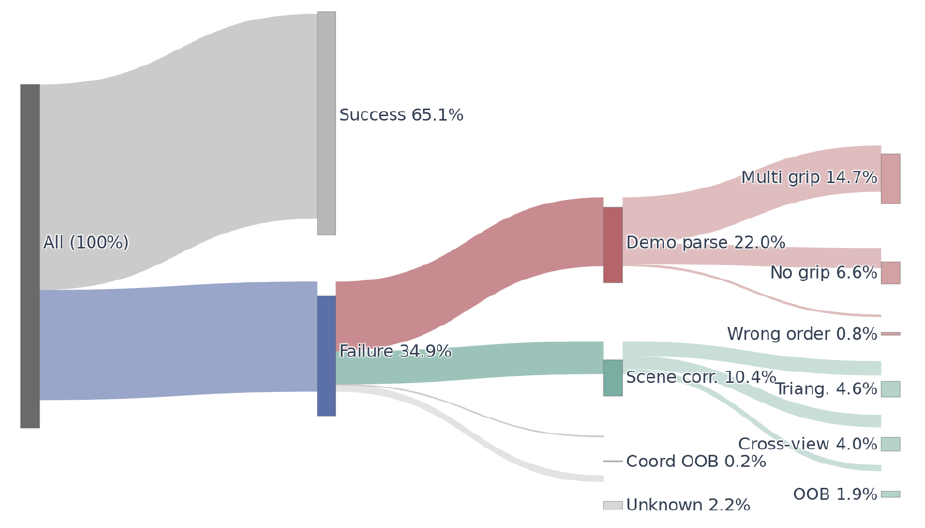}

    \caption{
        \textbf{Failure-mode breakdown in RoboLab.}
        Sankey diagram showing the distribution of warping outcomes across
        successful executions and major failure categories. \textit{Demo parsing errors} (22.0\%) are source-side: the DROID demonstration exhibits end-effector behavior incompatible with the target task, such as multiple or missing gripper open/close events or an incorrect grasp-release order, and is discarded before warping. \textit{Scene correspondence errors} (10.4\%) are target-side: triangulation failure, cross-view inconsistency, or out-of-bounds projection of the detected waypoints. These percentages describe warping outcomes, not task success rates, which are reported in Table~\ref{tab:robolab_trajectory_scaling}.}
    \label{fig:robolab_failure_mode}
\vspace{-1em}
\end{figure}

%% file: root.bbl
\begin{thebibliography}{10}
\providecommand{\url}[1]{#1}
\csname url@rmstyle\endcsname
\providecommand{\newblock}{\relax}
\providecommand{\bibinfo}[2]{#2}
\providecommand\BIBentrySTDinterwordspacing{\spaceskip=0pt\relax}
\providecommand\BIBentryALTinterwordstretchfactor{4}
\providecommand\BIBentryALTinterwordspacing{\spaceskip=\fontdimen2\font plus
\BIBentryALTinterwordstretchfactor\fontdimen3\font minus \fontdimen4\font\relax}
\providecommand\BIBforeignlanguage[2]{{%
\expandafter\ifx\csname l@#1\endcsname\relax
\typeout{** WARNING: IEEEtran.bst: No hyphenation pattern has been}%
\typeout{** loaded for the language `#1'. Using the pattern for}%
\typeout{** the default language instead.}%
\else
\language=\csname l@#1\endcsname
\fi
#2}}

\bibitem{lewis2020retrieval}
P.~Lewis, E.~Perez, A.~Piktus, F.~Petroni, V.~Karpukhin, N.~Goyal, H.~K{\"u}ttler, M.~Lewis, W.-t. Yih, T.~Rockt{\"a}schel, \emph{et~al.}, ``Retrieval-augmented generation for knowledge-intensive nlp tasks,'' \emph{Advances in neural information processing systems}, vol.~33, pp. 9459--9474, 2020.

\bibitem{memmel2025strap}
M.~Memmel, J.~Berg, B.~Chen, A.~Gupta, and J.~Francis, ``Strap: Robot sub-trajectory retrieval for augmented policy learning,'' in \emph{The Thirteenth International Conference on Learning Representations}, 2025.

\bibitem{regent2025}
K.~Sridhar, S.~Dutta, D.~Jayaraman, and I.~Lee, ``Regent: A retrieval-augmented generalist agent that can act in-context in new environments,'' in \emph{International Conference on Learning Representations}, vol. 2025, 2025, pp. 39\,385--39\,414.

\bibitem{sridhar2025ricl}
------, ``Ricl: Adding in-context adaptability to pre-trained vision-language-action models,'' \emph{arXiv preprint arXiv:2508.02062}, 2025.

\bibitem{jang2026ra}
S.~Jang, M.~Jeon, M.~Kim, S.~Choi, D.~Kim, and H.~Yu, ``Ra-vla: Retrieval-augmented vla for test-time adaptation,'' \emph{arXiv preprint arXiv:2608.25585}, 2026.

\bibitem{liu2026vls}
S.~Liu, I.~S. Singh, Y.~Xu, J.~Duan, and R.~Krishna, ``Vls: Steering pretrained robot policies via vision-language models,'' \emph{arXiv preprint arXiv:2602.03973}, 2026.

\bibitem{omniguide2026}
Y.~Song, L.~Le, Y.-H. Park, J.~Wang, J.~Shi, L.~Liu, J.~Gu, E.~Eaton, D.~Jayaraman, and K.~Daniilidis, ``Omniguide: Universal guidance fields for enhancing generalist robot policies,'' \emph{arXiv preprint arXiv:2603.10052}, 2026.

\bibitem{du2023demodiffusion}
S.~Park, H.~Bharadhwaj, and S.~Tulsiani, ``Demodiffusion: One-shot human imitation using pre-trained diffusion policy,'' \emph{arXiv preprint arXiv:2506.20668}, 2025.

\bibitem{molmoact2025}
J.~Lee, J.~Duan, H.~Fang, Y.~Deng, S.~Liu, B.~Li, B.~Fang, J.~Zhang, Y.~R. Wang, S.~Lee, \emph{et~al.}, ``Molmoact: Action reasoning models that can reason in space,'' \emph{arXiv preprint arXiv:2508.07917}, 2025.

\bibitem{liang2026tether}
W.~Liang, S.~Wang, H.-J. Wang, O.~Bastani, Y.~J. Ma, and D.~Jayaraman, ``Tether: Autonomous functional play with correspondence-driven trajectory warping,'' \emph{arXiv preprint arXiv:2603.03278}, 2026.

\bibitem{zhang2025peek}
J.~Zhang, M.~Memmel, K.~Kim, D.~Fox, J.~Thomason, F.~Ramos, E.~B{\i}y{\i}k, A.~Gupta, and A.~Li, ``Peek: Guiding and minimal image representations for zero-shot generalization of robot manipulation policies,'' \emph{arXiv preprint arXiv:2509.18282}, 2025.

\bibitem{shi2025zeromimic}
J.~Shi, Z.~Zhao, T.~Wang, I.~Pedroza, A.~Luo, J.~Wang, J.~Ma, and D.~Jayaraman, ``Zeromimic: Distilling robotic manipulation skills from web videos,'' in \emph{2025 IEEE International Conference on Robotics and Automation (ICRA)}.\hskip 1em plus 0.5em minus 0.4em\relax IEEE, 2025, pp. 16\,939--16\,947.

\bibitem{li2025novaflow}
H.~Li, L.~Sun, Y.~Hu, D.~Ta, J.~Barry, G.~Konidaris, and J.~Fu, ``Novaflow: Zero-shot manipulation via actionable flow from generated videos,'' \emph{arXiv preprint arXiv:2510.08568}, 2025.

\bibitem{intelligence2025pi05}
\BIBentryALTinterwordspacing
K.~Black, N.~Brown, J.~Darpinian, K.~Dhabalia, D.~Driess, A.~Esmail, M.~R. Equi, C.~Finn, N.~Fusai, M.~Y. Galliker, D.~Ghosh, L.~Groom, K.~Hausman, b.~ichter, S.~Jakubczak, T.~Jones, L.~Ke, D.~LeBlanc, S.~Levine, A.~Li-Bell, M.~Mothukuri, S.~Nair, K.~Pertsch, A.~Z. Ren, L.~X. Shi, L.~Smith, J.~T. Springenberg, K.~Stachowicz, J.~Tanner, Q.~Vuong, H.~Walke, A.~Walling, H.~Wang, L.~Yu, and U.~Zhilinsky, ``$\pi_{0.5}$: a vision-language-action model with open-world generalization,'' in \emph{Proceedings of The 9th Conference on Robot Learning}, ser. Proceedings of Machine Learning Research, J.~Lim, S.~Song, and H.-W. Park, Eds., vol. 305.\hskip 1em plus 0.5em minus 0.4em\relax PMLR, 27--30 Sep 2025, pp. 17--40. [Online]. Available: \url{https://proceedings.mlr.press/v305/black25a.html}
\BIBentrySTDinterwordspacing

\bibitem{khazatsky2024droid}
A.~Khazatsky, K.~Pertsch, S.~Nair, A.~Balakrishna, S.~Dasari, S.~Karamcheti, S.~Nasiriany, M.~K. Srirama, L.~Y. Chen, K.~Ellis, \emph{et~al.}, ``Droid: A large-scale in-the-wild robot manipulation dataset,'' \emph{arXiv preprint arXiv:2403.12945}, 2024.

\bibitem{liu2024grounding}
S.~Liu, Z.~Zeng, T.~Ren, F.~Li, H.~Zhang, J.~Yang, Q.~Jiang, C.~Li, J.~Yang, H.~Su, \emph{et~al.}, ``Grounding dino: Marrying dino with grounded pre-training for open-set object detection,'' in \emph{European conference on computer vision}.\hskip 1em plus 0.5em minus 0.4em\relax Springer, 2024, pp. 38--55.

\bibitem{senin2008dynamic}
P.~Senin, ``Dynamic time warping algorithm review,'' \emph{Information and Computer Science Department University of Hawaii at Manoa Honolulu, USA}, vol. 855, no. 1-23, p.~40, 2008.

\bibitem{yang2026robolab}
J.~X. Yang, R.~Dagli, A.~Zook, H.~Hadfield, A.~Goyal, S.~Birchfield, F.~Ramos, and J.~Tremblay, ``Robolab: A high-fidelity simulation benchmark for analysis of task generalist policies,'' \emph{arXiv preprint arXiv:2604.09860}, 2026.

\bibitem{yu2019meta}
T.~Yu, D.~Quillen, Z.~He, R.~Julian, K.~Hausman, C.~Finn, and S.~Levine, ``Meta-world: A benchmark and evaluation for multi-task and meta reinforcement learning,'' in \emph{Conference on robot learning}.\hskip 1em plus 0.5em minus 0.4em\relax PMLR, 2020, pp. 1094--1100.

\bibitem{robocasa2024}
S.~Nasiriany, A.~Maddukuri, L.~Zhang, A.~Parikh, A.~Lo, A.~Joshi, A.~Mandlekar, and Y.~Zhu, ``Robocasa: Large-scale simulation of everyday tasks for generalist robots,'' \emph{arXiv preprint arXiv:2406.02523}, 2024.

\bibitem{nvidia2025gr00tn1openfoundation}
\BIBentryALTinterwordspacing
NVIDIA, J.~Bjorck, F.~Castañeda, N.~Cherniadev, X.~Da, R.~Ding, L.~J. Fan, Y.~Fang, D.~Fox, F.~Hu, S.~Huang, J.~Jang, Z.~Jiang, J.~Kautz, K.~Kundalia, L.~Lao, Z.~Li, Z.~Lin, K.~Lin, G.~Liu, E.~Llontop, L.~Magne, A.~Mandlekar, A.~Narayan, S.~Nasiriany, S.~Reed, Y.~L. Tan, G.~Wang, Z.~Wang, J.~Wang, Q.~Wang, J.~Xiang, Y.~Xie, Y.~Xu, Z.~Xu, S.~Ye, Z.~Yu, A.~Zhang, H.~Zhang, Y.~Zhao, R.~Zheng, and Y.~Zhu, ``Gr00t n1: An open foundation model for generalist humanoid robots,'' 2025. [Online]. Available: \url{https://arxiv.org/abs/2503.14734}
\BIBentrySTDinterwordspacing

\bibitem{balasubramanian2011robust}
S.~Balasubramanian, A.~Melendez-Calderon, and E.~Burdet, ``A robust and sensitive metric for quantifying movement smoothness,'' \emph{IEEE transactions on biomedical engineering}, vol.~59, no.~8, pp. 2126--2136, 2011.

\end{thebibliography}
